\documentclass[letterpaper, 10 pt, conference]{ieeeconf}  

\IEEEoverridecommandlockouts                              

\usepackage[hidelinks, colorlinks=true, linkcolor=black, citecolor=green]{hyperref}
\usepackage{graphics} 
\usepackage{epsfig} 
\usepackage{times} 
\usepackage{amsmath} 
\usepackage{amssymb}  
\usepackage{verbatim} 
\usepackage{pifont}
\usepackage{blindtext}
\usepackage{xcolor} 
\usepackage{cuted}      
\usepackage{capt-of}    
\usepackage{graphicx}   
\usepackage{multirow}
\usepackage{booktabs}
\usepackage{url}
\usepackage{algorithm}
\usepackage{cite}
\usepackage{algpseudocode}

\algnewcommand\Param{\item[\textbf{Parameter:}]}
\title{\LARGE \bf
Glance-Say: Multimodal Human-Robot Collaboration and Intent Recognition via Sticky Glance
}

\author{Yuzhi Lai$^{1}$,~Shenghai Yuan$^{2}$,~Peizheng Li$^{1,3}$,~Benjamin Kiefer$^{1}$ and Andreas Zell$^{1}$
\thanks{Corresponding Author: \textbf{Andreas Zell}. This work was supported in part by Meta, and we acknowledge their contribution of Meta Glasses for this research.}
\thanks{$^{1}$University of Tuebingen,  Geschwister-Scholl-Platz, 72074 Germany, 
        {\tt\small \{name.surname\}@uni-tuebingen.de}.}%
\thanks{$^{2}$Nanyang Technological University, 50 Nanyang Avenue, Singapore 639798, 
        {\tt\small shyuan@ntu.edu.sg}.}%
        \thanks{$^{3}$
        {\tt\small peizheng.li@mercedes-benz.com}.}%
}

\begin{document}

\bstctlcite{IEEEexample:BSTcontrol}

\maketitle
\thispagestyle{empty}
\pagestyle{empty}

\begin{abstract}

Gaze and speech are promising interaction modalities for individuals with motor impairments, yet robust intent recognition in multi-object environments remains challenging due to micro-saccades, semantic ambiguity, and viewpoint changes. This paper presents a multimodal interaction framework for assistive robotic manipulation. We propose a sticky-glance algorithm that stabilizes gaze-based intent by jointly accumulating geometric distance and directional evidence, enabling robust real-time target selection and switching. We further introduce Glance-Say, a gaze–speech interaction paradigm in which gaze specifies objects and speech specifies actions, together with a continuous shared-control scheme that provides high-readiness robot motion and human-in-the-loop feedback. Experiments demonstrate a tracking rate of 0.92 for moving targets, selection accuracy of 0.97 for static targets, and reduced task duration. These results indicate improved robustness, efficiency, and usability over representative interaction paradigms.


\end{abstract}

\section{INTRODUCTION}
Gaze provides a direct, low-effort channel to express intent, and has long served as an input modality in robotic manipulation \cite{8593766, lai2025famhrifoundationmodelassistedmultimodal}. This capability is especially important in assistive robotics for users with severe motor impairments, who often retain intact cognition and voluntary eye movements \cite{c24}. In this setting, a practical gaze interface must go beyond gaze-space smoothing: it must reliably infer intent in real time and perform object-level grounding in cluttered scenes, 
while supporting safe and efficient human-in-the-loop control.

\textbf{Existing} gaze-based pipelines are biased toward either temporal persistence or statistical smoothing, which makes them brittle in realistic, dynamic environments \cite{10.1145/3656376}. Prolonged-fixation designs trade responsiveness for stability and can still break under jitter and transient glances \cite{tay2025intent}. Distribution-based approaches reduce noise but often ignore object-centric geometric relationships of gaze trajectories and explicit temporal constraints \cite{10806588}. Probabilistic models improve interpretability yet typically rely on delicate parameter tuning and manually specified transitions that do not transfer cleanly across users and scenes \cite{10.3389/fnbot.2021.647930, 8593766}. Learning-based solutions require extensive labeled data and often generalize poorly when users or object layouts change.  Beyond intent recognition, many gaze-driven robotic systems still operate with a discrete target-pose triggering controller, offering limited continuous feedback during intent formation and thus constraining responsiveness in assistive use \cite{lai2025famhrifoundationmodelassistedmultimodal, tay2025intent}. 

The core \textbf{challenge} is that robotic manipulation requires object-level geometric intent grounding rather than gaze-space smoothing. This challenge involves a tripartite coupling of (i) noisy, nonstationary gaze trajectories (micro-saccades, head motion), (ii) multi-object dynamics and viewpoint changes that make gaze-to-object grounding ambiguous, and (iii) control requirements that demand both stability (avoid unintended selection) and immediacy (avoid long dwell times). Solving only the perception side is insufficient if the downstream interaction remains discrete and non-responsive; conversely, improving control without robust object-level intent grounding leads to unsafe or frustrating operation. 

To address these challenges, we propose an object-centric gaze grounding and multimodal interaction framework that stabilizes intent directly in geometric space rather than solely in gaze space. Instead of relying on prolonged fixation or probabilistic belief propagation and transition probability, we accumulate geometric and temporal evidence by jointly modeling distance and directional trends of gaze motion. This allows intent to ``stick" to object regions even under short glances and dynamic conditions, while also enabling seamless target switching. Building on this stabilized intent signal, we further introduce a continuous shared controller and interaction paradigm that couples gaze-based object selection with speech-based action specification. It synthesizes (i) a virtual target anchored to grounded objects and (ii) a gaze-conditioned velocity field. Unlike conventional target-pose triggering, our system provides continuous motion feedback and slow-following during intent formation, enabling more responsive and efficient gaze-driven HRI.

\begin{figure}[tp]
      \centering
      \includegraphics[width=0.40\textwidth]{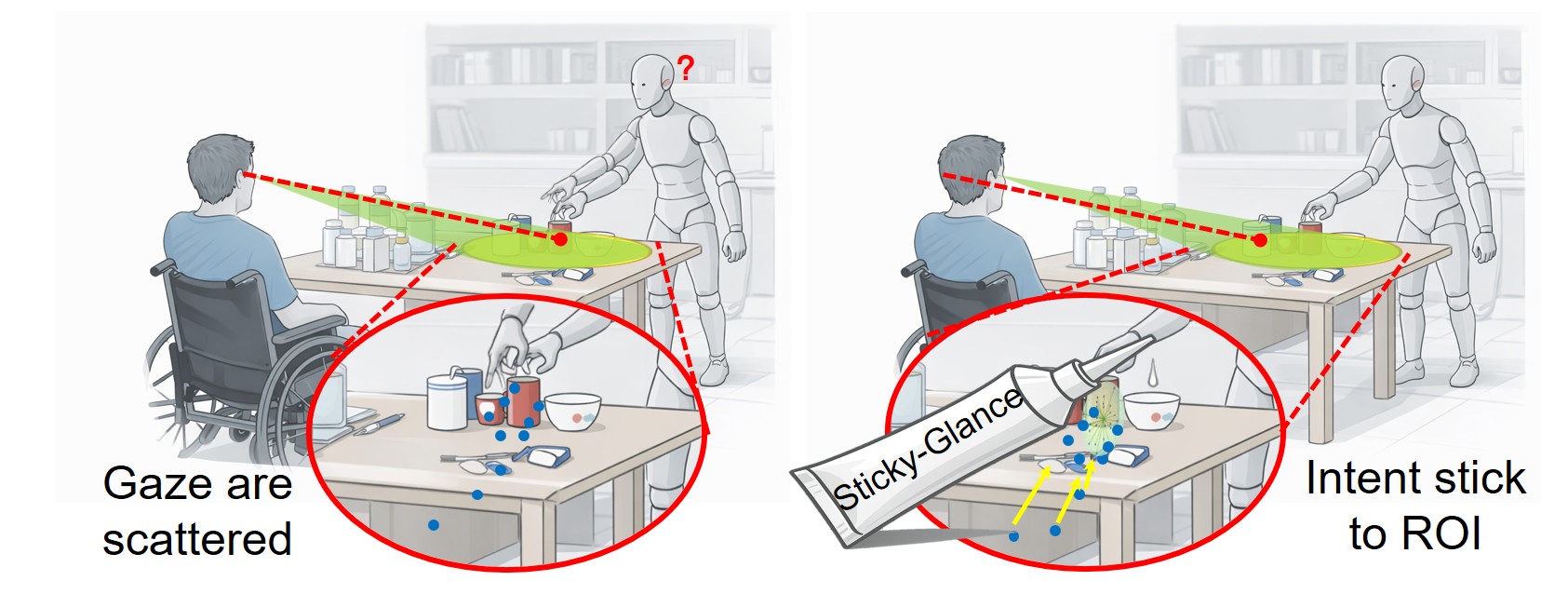}
     \vspace{-5pt}
      \caption{ Our proposed approach anchors scattered gaze points to intent objects, enabling interaction with a short glance.}
      \label{f1}
      \vspace{-15pt}
\end{figure} %

Our main contributions are summarized below: 
\begin{itemize}

\item 
We propose a gaze grounding and intent stabilization module that maps noisy gaze into an object-centric gaze-motion evidence field, maintaining robust target assignment and switching under micro-saccades and dynamic object motion without prolonged fixation, per-task fine-tuning or initialization.



\item 

We introduce a confidence-aware continuous control strategy that switches between standby behavior and rapid attraction, improving motion predictability and reducing task duration.

\item 


We develop Glance-Say, a multimodal interaction framework that replaces GUI-based or fixation-only interaction with continuous gaze grounding and speech-based action specification. It enables robust viewpoint alignment between robot and user and real-time feedback while demonstrating improved success rate and usability, and reduced cognitive load

\end{itemize}

\section{Related Work}

The goal of gaze-based intent recognition is to infer which object a user intends to interact with based on gaze signals. Existing approaches can be broadly categorized into model-based, learning-based, distribution-based, and fixation-based. 
Model-based approaches formulate intent prediction as a sequence modeling or classification problem using Hidden Markov Models (HMMs) \cite{10.3389/fnbot.2021.647930, 8593766}, Support
Vector Machines (SVMs) \cite{7451737}, or Partially Observable Markov Decision Processes (POMDPs) \cite{admoni2016predicting}. Some researchers also utilized learning-based methods such as Recurrent Neural Networks (RNNs) \cite{wang2020toward} or Long Short-Term Memory (LSTM) \cite{gonzalez2019perceptually}.
These methods model gaze sequences over temporal windows, but their performance depends heavily on window size, labeled data, and parameter tuning. In practice, performance varies across users and environments, limiting robustness and real-world scalability. TAGMM \cite{10023502} models gaze dynamics using a Kalman filter with an artificial attractor field, assuming that gaze trajectories follow a physically driven acceleration toward the intended target. However, human gaze behavior does not strictly obey such deterministic attractor dynamics. Moreover, the model requires 800 ms of gaze history for initialization and parameter stabilization, introducing a non-negligible latency in the interaction system.
Distribution-based methods estimate intent by aggregating gaze samples over time, such as through clustering \cite{10806588}, weighted trajectory accumulation \cite{lai2025famhrifoundationmodelassistedmultimodal}, or spatial probability modeling \cite{shi2024casualgaze}. Although they reduce reliance on labeled data, they operate primarily in gaze space and do not explicitly model motion trends or geometric constraints between gaze trajectories and objects. The most widely adopted strategies rely on fixation or dwell-time thresholds \cite{tay2025intent,c7}, where an object is selected only when a sufficient number of gaze samples fall within its region. While simple to implement, these approaches require prolonged fixation and are sensitive to natural micro-saccades. In static scenes, involuntary gaze jitter may cause gaze points to drift outside object boundaries, leading to selection failure. In dynamic scenes, tracking performance further degrades because sustained fixation on moving objects becomes difficult.
A degenerate case of this strategy reduces selection to the last gaze point with a k-Nearest Neighbor-based (kNN) approach \cite{11037823   }. 
However, even a single noise sample can lead to an incorrect prediction. Gaze-based intent recognition should move beyond smoothing gaze samples or estimating static gaze-object associations within a temporal window. Instead, it should model the instantaneous object-centric geometry of gaze motion, capturing the relationship between the gaze trajectory and each candidate object.


Existing gaze-based robotic interaction strategies can be broadly categorized into interface-centric control \cite{11037823, c7, c24}, and multimodal interaction \cite{lai2025famhrifoundationmodelassistedmultimodal}.
Interface-centric approaches rely on gaze-controlled graphical user interfaces (GUIs). 
FreeView \cite{c24} introduces a virtual interface anchored around the robot end-effector, where users operate the manipulator by fixating on interactive elements positioned near the tool. Users must continuously attend to interface elements, resulting in long command duration, high cognitive load, and considerable learning cost \cite{11127558}. To reduce complexity, MR-GUI \cite{c7} simplifies the interaction by allowing users to first select an object through gaze and then choose the desired action via a GUI. Although this design shortens command duration compared to full teleoperation interfaces, it still relies on GUI elements, constraining interaction to predefined interface locations. In contrast, GlanceGaze \cite{tay2025intent} removes the explicit GUI by leveraging a vision-language model (VLM) to infer the intended action directly from gaze and scene context. In \cite{wang2023gaze}, similarly, researchers utilize gaze and LSTM to infer action primitives and target pose. However, these action inferences are unreliable in ambiguous scenarios (e.g., distinguishing between “put” and “pour”), affecting task stability and success rate. Multimodal approaches, such as FAM-HRI \cite{lai2025famhrifoundationmodelassistedmultimodal} and HIP-HRI \cite{11037823}, combine gaze for object selection with speech for action specification. These methods integrate the advantages of both modalities, reducing ambiguity and lowering cognitive load.
However, existing multimodal systems are still mainly designed around discrete target selection and target-pose triggering. FAM-HRI \cite{lai2025famhrifoundationmodelassistedmultimodal} estimates the intended object using weighted gaze clustering over a temporal window. Although this improves robustness over kNN- or Fixation-based methods, it does not explicitly and continuously infer the object-centric geometric relationship between gaze motion and candidate objects. Moreover, as a result of discrete intent inference, existing systems trigger robot motion only after intent confirmation and determination of the target position, providing limited real-time feedback. Additionally, prior systems\cite{11037823, tay2025intent, wang2023gaze} mainly focus on human-view object selection and give limited attention to viewpoint-induced occlusion and multi-perspective object alignment between the user and the robot.


\section{Methodology}
\subsection{Problem Formulation}
\label{Problem Formulation}

\begin{figure*}[tb]
    \vspace{5pt}
      \centering
      \includegraphics[width=0.85\textwidth]{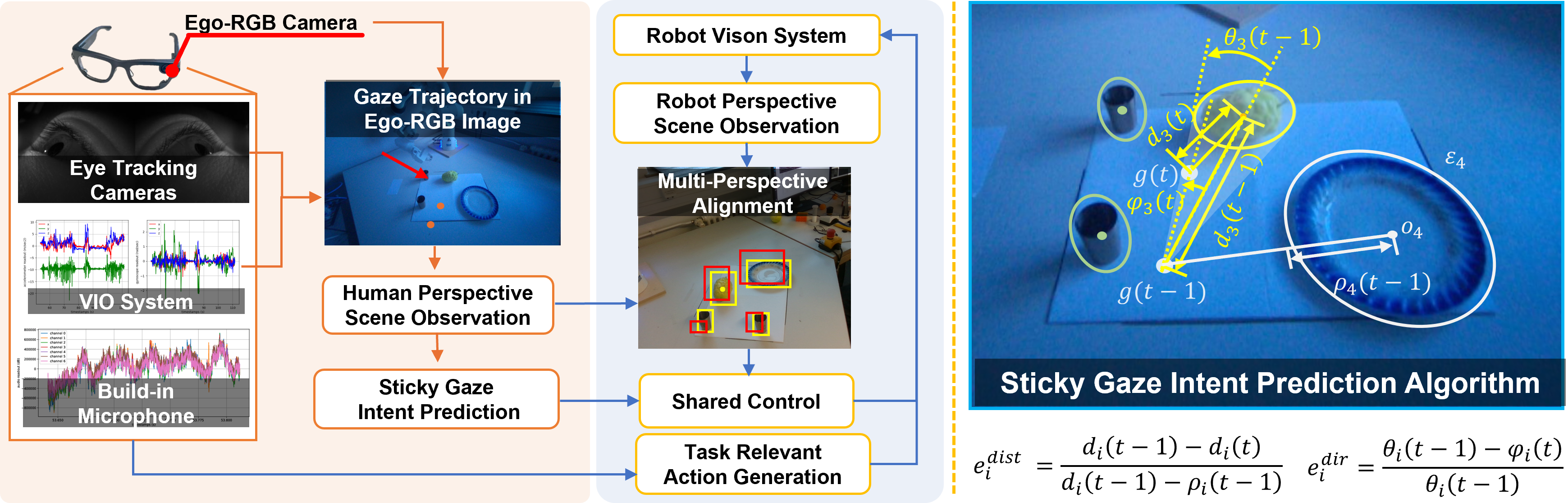}
      \caption{ Our proposed Glance-Say system collects data through Meta ARIA glasses, then performs asynchronous off-device inference. Data is transmitted via Wi-Fi. Sticky-Glance Intent Prediction Algorithm simultaneously models the directional and distance trend of users' gaze.
      }
      \label{f2}
      \vspace{-15pt}
\end{figure*}

Our goal is to infer object-level user intent from gaze points $g(t) \in \mathbb{R}^2$ over time $t$ in a 3D workspace with objects $\mathcal{Z}_{{i}} \in \mathbb{R}^3$. The objective is to reliably infer the intended object $\mathcal{Z}_{i^{\star}} \in \mathbb{R}^3$. 
Our system is shown in Fig. \ref{f2} (left).


\subsection{Human Perspective Perception}
\label{Human View Perception}

\subsubsection{Gaze Trajectory Projection}
\label{Gaze Trajectory Projection}
We use the eye-tracking cameras of Meta ARIA glasses to obtain per-frame gaze points in the ego-image \cite{11052289}. The built-in VIO system is utilized to estimate camera poses. The gaze point from the previous frame $g(t-1)$ is then reprojected into the current frame of the ego-image to obtain the gaze trajectory with the current gaze point
$g(t)$ as shown in Fig. \ref{f2} (right). 
\subsubsection{Human Perspective Scene Observation}
\label{Human Perspective Scene Observation}
 A fine-tuned YOLO26 \cite{yolo26e_ultralytics} is used for object detection and segmentation, and ByteTrack \cite{zhang2022bytetrack} is employed for multi-object tracking in the ego-RGB images. To handle object shape variation robustly, we represent each object by the outer ellipse of its mask in addition to its bounding box.

\subsection{Sticky-Glance Intent Prediction Algorithm}
\label{Sticky-Glance Intent Prediction Algorithm}
The goal of the sticky-glance algorithm is to construct an object-centric confidence field in which candidate objects attracts or repels gaze motion, enabling object-centric geometric intent grounding rather than gaze-space smoothing or static gaze-object associations.
As shown in Fig. \ref{f2} (right), each candidate object is represented by an elliptical region $\varepsilon_i$ centered at $o_i$. 
As shown by the light blue cone, we further define a tangent cone for each object. This cone is formed by the two tangent lines from the previous gaze point $g(t-1)$ to $\varepsilon_i$. 

We first define the distance between the gaze point $g(t)$ and ellipse center $o_i$ as: $d_i(t) = \left \| g(t)-o_i \right \|_2$
, and $\rho_i(t)$ denotes the distance from the ellipse center $o_i$ to the ellipse boundary along the line connecting $o_i$ and $g(t)$. Based on these, we further define a distance-evidence term:
\begin{align}
e^{\text{dist}}_i(t)=\text{clip}(\frac{d_i(t-1)-d_i(t)}{d_i(t-1)-\rho_i(t-1)}, \ -1, \ 1)\label{edist}   
\end{align}
We define $e^{\text{dist}}_i (t)$ in Eq. \ref{edist} to measure the normalized distance trend between the gaze trajectory and the elliptical region $\varepsilon_i$. Positive evidence is generated when the gaze moves closer to $\varepsilon_i$, while negative evidence is produced when it moves away. The boundary distance $d_i(t-1)-\rho_i(t-1)$ ensures $e_i^{\text{dist}}(t)$ reaches 1 when the gaze enters $\varepsilon_i$. After clipping, the evidence reaches -1 when the gaze moves away by twice the boundary distance, preventing excessive negative updates.

\begin{equation}
  e_i^{dir}(t)=\text{clip}(\frac{\theta_i(t-1)-\varphi_i(t)}{\theta_i(t-1)},\ -1, \ 1)\label{edir}
\end{equation}

By leveraging the tangent cone, we define the directional trend of the gaze in Eq.\ref{edir}. The cone represents the feasible angular region within which gaze motion would approach the object. Here, $\theta_i(t-1)$ denotes the angular boundary of the tangent cone formed by the vector $\overrightarrow{g(t-1)o_i}$ and the tangent line, while $\varphi_i(t)$ denotes the angle between the gaze displacement $\overrightarrow{g(t-1)g(t)}$ and $\overrightarrow{g(t-1)o_i}$. The evidence reaches 1 when $\varphi_i(t)=0$, indicating that the gaze moves directly toward the ellipse center. 
When $2\theta_i (t-1) \le \varphi_i(t)$, the gaze strongly deviates from the ellipse center, and the clipped evidence reaches -1.



We define $I_i(t)$ as the number of intersection points between the gaze displacement $\overrightarrow{g(t-1)g(t)}$ and $\varepsilon_i$. Our Sticky-Glance maintains a confidence score for each candidate object and updates it using distance and directional evidence. As shown in Alg. \ref{alg:glance_short}, we first compute $e_i^{\text{dist}}$ and $e_i^{\text{dir}}$ from the gaze motion. The evidence is then adjusted according to the intersection between the gaze trajectory and the object region: entering the object provides positive evidence, while leaving or passing through the object produces negative evidence. When the gaze remains inside the object with a small displacement, it is treated as a fixation and assigned positive distance evidence. When the gaze displacement exceeds the fixation threshold, the motion is treated as an intra-object gaze shift. In this case, we only keep the distance evidence to reflect whether the gaze is moving toward or away from the center $o_i$ ($\rho_i=0$). This is particularly useful when objects overlap each other. The final evidence $e_i$ is integrated into the confidence score with $\Delta t$. An object is selected as an intended target once its confidence exceeds $c_{\min}$. 
The integral update of confidence introduces a sticky effect, allowing intent to remain anchored to the object once established. As the confidence of the current target decreases and that of a new target increases, the intended object is switched through a gradual confidence transition rather than a predefined transition probability. This mechanism filters transient micro-saccades while enabling seamless intent switching. Multiple intents can exist simultaneously, please refer to Sec. \ref{Human-Robot Interaction}. 

\begin{figure}[t]
\begin{algorithm}[H]
\caption{Sticky-Glance Intent Prediction}
\label{alg:glance_short}
\begin{algorithmic}[1]

\Require Objects $\varepsilon_{i=1}^{N}$, gaze $\{g(t)\}_{t=0}^{T}$
\Ensure Confidence $\mathbf{c}(t, i)\in[0,1]^N$, Union of intent $\mathcal{I}^{\star}(t)$
\Param{Fixation threshold $\tau$, Step $\Delta t$, Threshold $c_{min}$, Number of Intersections $I_i(t)$},

\For{$i=1$ to $N$}
\State $e^{\text{dir}}_i\gets$Eq. \ref{edir} , $e^{\text{dist}}_i\gets$Eq. \ref{edist}

\If{$I_i(t)=1\land d_i(t-1) > \rho_i$} 
    \State $e^{\text{dir}}_i, e^{\text{dist}}_i\gets 1$ \Comment{Gaze entering object}
\ElsIf{$I_i(t)=1\land d_i(t-1) < \rho_i$}
    \State $e^{\text{dir}}_i, e^{\text{dist}}_i\gets -1$ \Comment{Gaze leaving object}
\ElsIf{$I_i(t)=2$}
    \State $e^{\text{dir}}_i, e^{\text{dist}}_i\gets -1$ \Comment{Gaze passing object}
\EndIf

    \If{$ d_i(t-1)\le \rho_i \land \ d_i(t)\le \rho_i \land \Delta g(t) < \tau$}
      \State $e^{\text{dist}}_i\gets 1$,$e^{\text{dir}}_i\gets 0$  \Comment{Fixation on object}
       \ElsIf{$ d_i(t-1)\le \rho_i \land \ d_i(t)\le \rho_i \land \Delta g(t) \ge \tau$} 
       \State $\rho_i\gets0$,$e^{\text{dir}}_i\gets 0$, $e^{\text{dist}}_i\gets $Eq. \ref{edist}  \Comment{Intra gaze shift}
      \EndIf

      \State $e_i \gets e^{\text{dist}}_i + e^{\text{dir}}_i$
      \State $\mathbf{c}(t, i)\gets \mathrm{clip} (\mathbf{c}(t-1, i)+\Delta t\cdot e_i,\ 0,\ 1)$
  
 \If{$\mathbf{c}(t, i) \ge c_{\min}$}
   \State $\mathcal{I}^{\star}(t) \gets \mathcal{I}^{\star}(t) \cup \{i \}$ \Comment{Intent update}
\Else
   \State \textbf{continue} \Comment{No selection for object $i$}
 \EndIf
\EndFor
\end{algorithmic}
  
\end{algorithm}
\vspace{-25pt}
\end{figure}

\vspace{-3pt}
\subsection{Robot Perspective Perception}

\subsubsection{Object-Level Point cloud Association}
\label{Object Level Pointcloud Association}

To avoid information loss from a single viewpoint, we construct complete object-level point clouds. 
Inspired by prior work \cite{10128836}, Fig.~\ref{association} shows our pipeline for object-level association and alignment. The system explores the workspace, aligning captured point clouds through Iterative Closest Point (ICP) registration. 
Isolation Forest is applied to remove the background and to ensure a smooth object geometry $\mathcal{Z}_i$.    
We then construct a topology graph, where each node represents an object with its semantic label (from Sec. \ref{Human Perspective Scene Observation}), 3D position, and pre-grasping pose \cite{fang2023anygrasp} (above the object). Edges encode spatial relations: ``\textit{on}" if vertical projections overlap, and ``\textit{in}" if horizontal projections are enclosed.

\subsubsection{Multi-Perspective Alignment}
\label{Multi-Perspective Alignment}

\begin{figure}[tp]
    \vspace{5pt}
      \centering
      \includegraphics[width=0.48\textwidth]{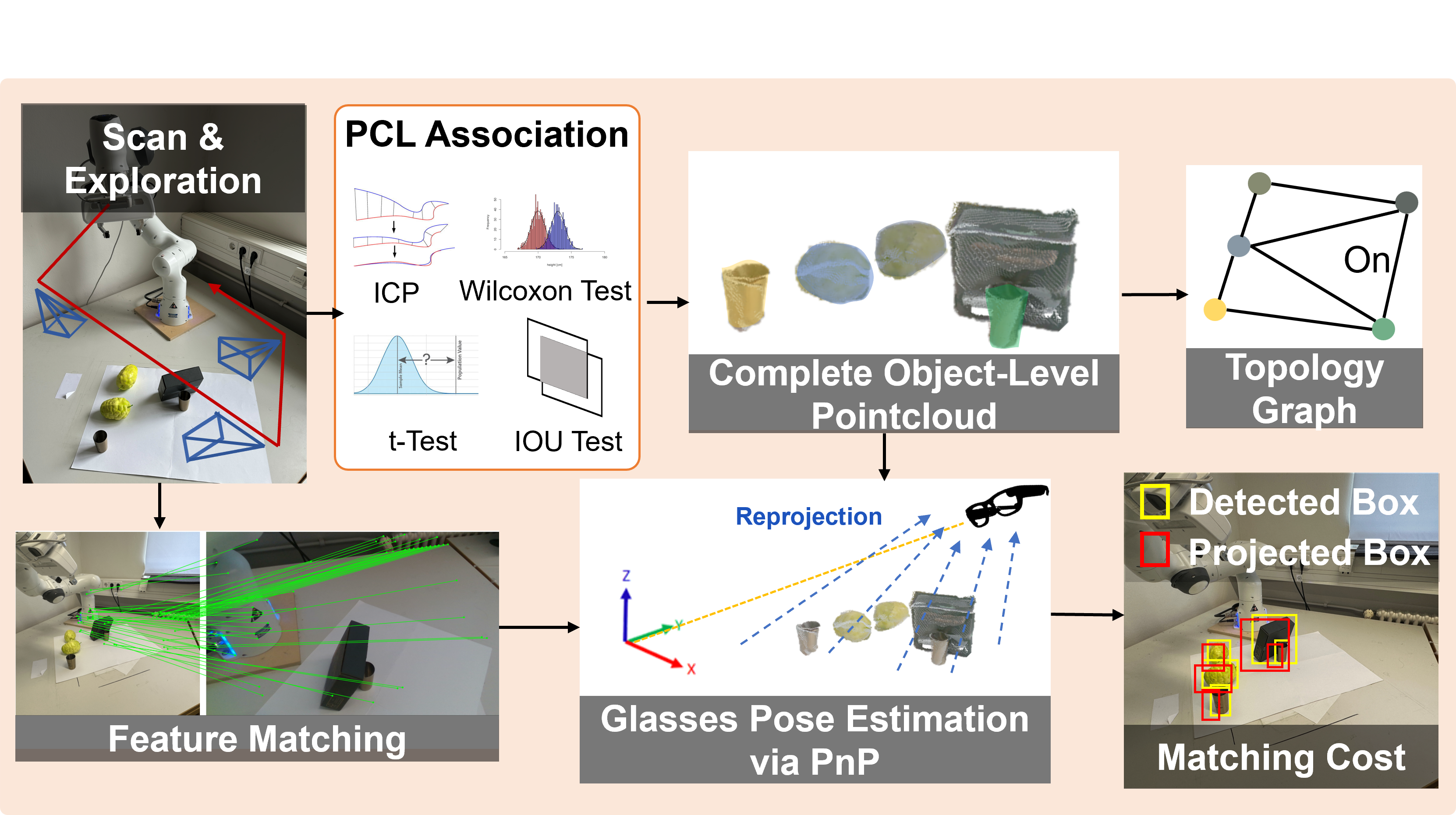} 
      \vspace{-15pt}
      \caption{Demonstration of the pipeline for object-level point cloud association and multi-perspective alignment.}
      \label{association}
      \vspace{-15   pt}
\end{figure}

To enable the robot to identify intent, multi-perspective alignment between the human and robot perspectives is required. Previous feature-based voting alignment \cite{lai2025famhrifoundationmodelassistedmultimodal, tay2025intent} often fails due to insufficient matching of object features and misdetection of ArUco markers. As shown in Fig. \ref{association}, we propose an optimal matching-based method. We first perform feature matching using LightGlue \cite{Lindenberger_2023_ICCV} and estimate the ego-RGB camera pose via Perspective-n-Point (PnP). Using this pose, 3D object point clouds (from Sec. \ref{Object Level Pointcloud Association}) are projected into the ego-RGB image to obtain normalized projected bounding boxes $b^p_m=[\frac{{x_m}}{W}, \frac{{y_m}}{H},  \frac{{w_m}}{W}, \frac{{h_m}}{H}]$, where $W$, $H$ represent the width and height of the ego-RGB image. With normalized detected boxes $b^d_n$ from Sec. \ref{Human Perspective Scene Observation},
We define the cost:
\begin{equation}
\vspace{-2   pt}
   L = \left \| b^p_m-b^d_n \right \| _2
   \vspace{-2pt}
\end{equation}
where $m$ and $n$ represent the index of the bounding box. 
The Hungarian algorithm is applied to obtain optimal object correspondence with minimal $L$. To ensure real-time performance, multi-view alignment is only performed at the start of interaction (first frame) and when new objects appear. For failure treatment, please refer to Sec. \ref{Human-Robot Interaction}.



\subsection{Continuous Shared Control and Interaction}
\label{Human-Robot Interaction}

Continuous robot response is critical for intuitive interaction, as delayed or unstable motion can degrade usability and safety. 
During the confidence updating and intent transitions, multiple candidate objects may temporarily exhibit high confidence. To ensure smooth and predictable behavior, we define a confidence-weighted virtual target as: 
\begin{equation}
  \mathcal{P}(t) =\frac{\sum_{ i\in \mathcal{I}^{\star}(t)}(\mathbf{c}(t, {i})\mathcal{P}_g^{i})}{\sum_{  i\in\mathcal{I}^{\star}(t)}\mathbf{c}(t, {i})}
    \label{Pt}
\end{equation}
  $\mathcal{P}_g^{i}$ denotes the pre-grasping position, which is decomposed from pre-grasp pose $\mathcal{T}_g^{i}$ (see Sec. \ref{Object Level Pointcloud Association}). We also define two control modes: a pre-command mode for continuous standby and a post-command mode for task execution.

Prior to the user providing an explicit task command through speech, the robot operates in pre-command mode. For the robot end-effector, we set the angular velocity to $\omega_{pre}(t)=0$ and define the linear velocity as: 
\begin{align}
v_{pre}(t) &= \underbrace{\frac{\sum_{ i\in \mathcal{I}^{\star}(t)}(\mathbf{c}(t, {i})v_{max})}{\left |  \mathcal{I}^{\star}(t)\right | }}_{v(t)}\sigma_e(t)u(t)
\end{align}
$v(t)$ is a confidence-weighted velocity that increases with confidence. $\sigma_e(t) = \text{clip}(\frac{\mathcal{P}_e(t)-\mathcal{P}(t)}{\Delta R},0,1)$ modulates the velocity based on the distance to the target, reducing speed when approaching a predefined safety distance $\Delta R$ to ensure smooth deceleration. The unit vector $u(t)= \frac{\mathcal{P}(t)-\mathcal{P}_e(t)}{\left \| \mathcal{P}(t)-\mathcal{P}_e(t) \right \| _2}$ defines the motion direction from the current end-effector position $\mathcal{P}_  e(t)$ to the virtual target.

Upon detecting a speech command, the system switches to post-command mode and commits to a single intended object. The index of the intended object $i_m$ is selected as $i_m(t)=\text{arg}\max_{{i\in\mathcal{I}^{\star}(t)}}c(t,{i}) $, which resolves the virtual target to a real target. In post-command mode, the robot moves towards the intended object with maximum velocity while ensuring a safe approach and then aligns its orientation with the pre-grasp orientation.
\begin{align}
     v_{post}(t)&= v_{max}\sigma_e(t, i_m)u(t,i_m) 
 \label{eq6}
\end{align}

To avoid interaction failures caused by intent recognition algorithms and multi-view matching, after reaching the target, the robot pauses and waits for user confirmation via speech. If the user confirms, the corresponding object is selected as the intended object $\mathcal{Z}_{i^{\star}}(t)$ and the robot proceeds with task execution. If the user rejects the selection, the robot moves to the object with the second highest confidence with Eq. \ref{eq6}, and repeats the confirmation process.  If the user rejects again, the robot returns to its initial pose, re-infers and waits for the next user intent. For ambiguous speech commands, the system repeatedly asks for clarification and postpones execution until an unambiguous command is received. The user can also issue a ``stop'' command at any time to immediately halt robot motion, empty the gripper if necessary, and return the robot to its initial pose. This confirmation and recovery mechanism ensures safe interaction while allowing direct correction of both intent and command ambiguity.


\subsection{Task Relevant Action Generation}
\label{Action Sequence Generator}

To enable real-time task planning from speech and glance, we build on the previous approaches \cite{8664854}. The speech command is encoded using a BGE model \cite{bge_m3} and matched with action primitives based on similarity. The primitives include \textit{pick, put, pour, swap, grasp}, \textit{push}, \textit{move} and \textit{rotate}.

The selected action is executed using a behavior tree planner integrated with the topology graph, enabling topology-aware manipulation. This allows the system to automatically generate intermediate actions, such as removing obstructing objects before manipulating the target, ensuring robust execution in complex scenes.

\section{Experiment Settings}
\label{Experiment Design}



We set $\Delta t=0.3$, $c_{\min}=0.3$, $\tau=50$ pixels, $v_{\max}=10~\mathrm{cm/s}$, and $\Delta R=5~\mathrm{cm}$, with all parameters selected by grid search on a separate tuning set. The robot end-effector is initialized above the objects to avoid collisions.

Human-view inputs are collected using Meta ARIA glasses, and the robot view is captured by an Intel RealSense D435i RGB-D camera. Manipulation is performed with a Franka Emika Panda 7-DoF robot. All multimodal processing runs on an NVIDIA RTX 4080 GPU, with models deployed using TensorRT. Glance-Say operates asynchronously, triggering multi-perspective alignment at interaction start and when new objects appear in less than 20 ms. The remaining components run in less than 18 ms, faster than the eye-tracking camera (10 FPS) and the ego-RGB camera (20 FPS).

As shown in Fig. \ref{experiment}, We design four scenarios:
    \begin{figure*}[htp]
      \centering
      \includegraphics[width=0.9\textwidth]{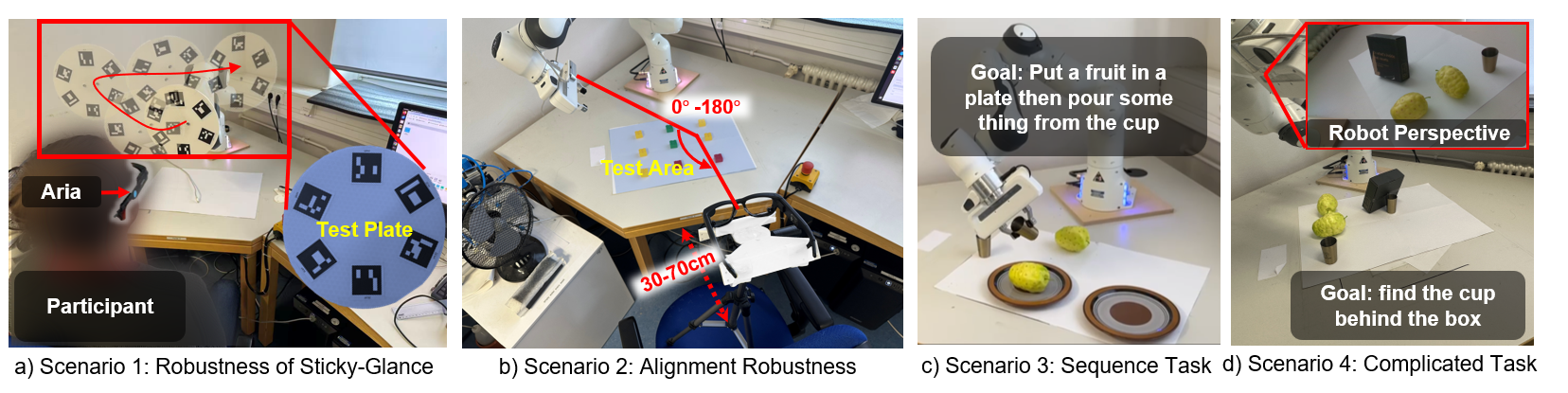}
     \vspace{-9pt}
      \caption{Overview of Experimental Scenarios. Scenario 1 evaluates the robustness of our intent confidence algorithm. Scenario 2 evaluates the robustness of multi-perspective alignment. Scenarios 3 and 4 assess real-world robotic task execution at increasing complexity. Sequence tasks involve combinations of simple actions such as pick, put, pour, and position swap. Complicated tasks require the robot to handle occlusion issues and the overlapping of objects.}
      \label{experiment}
      \vspace{-9pt}
\end{figure*} %

\begin{itemize}
\item Scenario 1 ($S_1$): Users wear ARIA glasses. For moving targets, we evaluate tracking rate, defined as the ratio of correctly identified intent frames to detected frames. For static targets, participants select specified targets using gaze, including intent switching. Selection accuracy is computed as the ratio of correct selections to total trials. Participants maintained an approximately 40 cm distance from the test plate to ensure accurate gaze points (error $\approx$ 0.4 cm).

\item Scenario 2 ($S_2$): The glasses are fixed at a representative user-like position to eliminate head motion. We evaluate object alignment under different distances and relative viewing angles between the glasses and the robot-mounted camera. For each condition, we conduct 100 independent trials, randomly selecting one target object per trial. Alignment is successful only if the matched object correspondence is correct. We report the number of successful trials out of 100. Nine visually similar Duplo blocks are arranged in a $3 \times 3$ grid to reduce object detection bias.

\item Scenarios 3 and 4 ($S_3$ and $S_4$): Users complete predefined manipulation tasks with different interaction methods. We report success rate, command duration, and task duration. Command duration measures user input time, while task duration measures the time from command start to robot execution completion.

\end{itemize}

We recruited 16 participants with upper-limb disabilities from multiple countries who were fluent in English. In scenarios $S_1$, $S_3$, and $S_4$, participants repeated our method and the baseline methods ten times. To minimize learning effects, the order of the scenarios and interaction methods was randomized across trials. 

\section{Discussion}

\subsection{Intent Recognition Robustness}

\begin{figure}[t]
      \centering
      \vspace{-5pt}
      \includegraphics[width=0.46\textwidth]{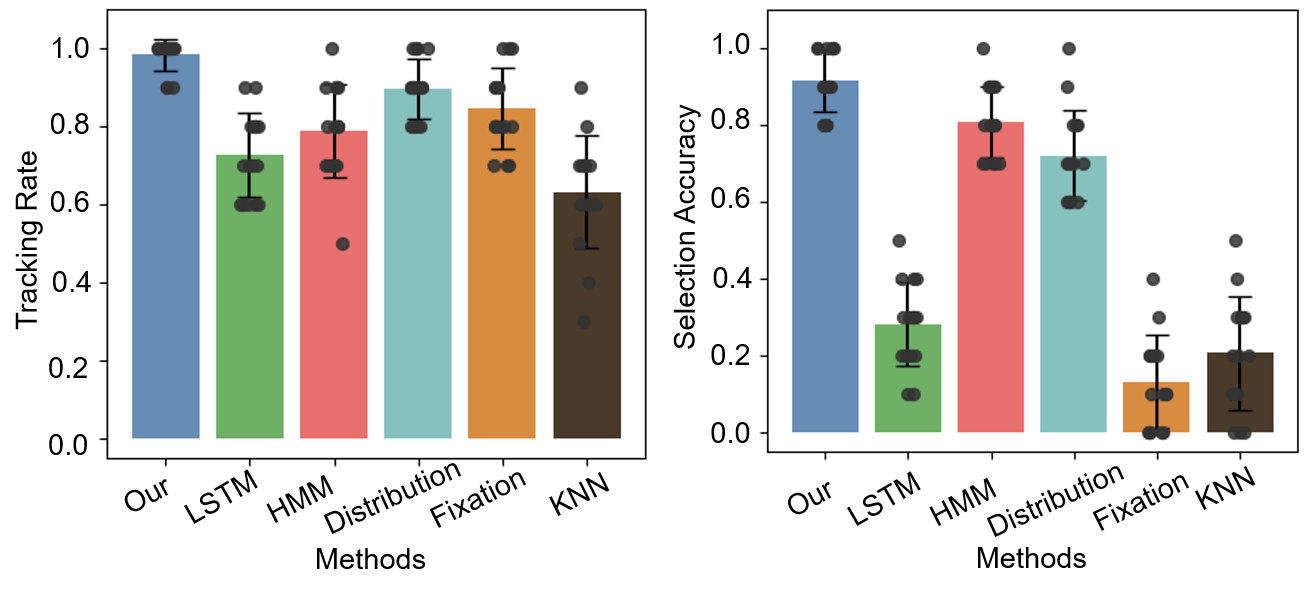} 
     \vspace{-10pt}
      \caption{Results for Selection Accuracy (right) and Tracking Rate (left).}
      \label{fig5}
\end{figure}

Tab. \ref{intent} and Fig. \ref{fig5} show the tracking rate under dynamic motion and the selection accuracy under static conditions in $S_1$ of different intent recognition baselines. To evaluate intent recognition robustness, we compare our method with state-of-the-art gaze-based intent recognition baselines, including kNN-based \cite{11037823}, fixation-based \cite{tay2025intent}, distribution-based \cite{10806588}, HMM-based \cite{8593766}, and LSTM-based \cite{gonzalez2019perceptually} approaches. Methods based on Kalman Filter that require initialization are not applicable \cite{10023502}. Additionally, aside from our method, only HMM-based methods perform object-centric geometric intent grounding; the remaining methods either infer static gaze-object relationships or simply smooth the gaze space.  

When targets move randomly, gaze often lags behind the object, causing baseline methods to lose target tracking and resulting in low tracking rates. 
Under static conditions, fixation-based \cite{tay2025intent} and kNN-based  \cite{11037823} methods suffer from natural micro-saccades, which can shift gaze outside the object region and reduce accuracy.  LSTM-based (and other learning-based ) \cite{wang2020toward, gonzalez2019perceptually} approaches infer intent based on all gaze points within a time window, leading to ambiguous intent recognition when intent transitions occur. 
HMM-based approach \cite{8593766} improves dynamic scenarios by introducing temporal belief propagation. However, the predefined goal-switching and transition probabilities are difficult to fine-tune. A small switching probability introduces strong inertia in belief updates, leading to delayed intent transitions in static scenarios, whereas a larger switching probability causes belief oscillation and instability in dynamic settings.
For each baseline, we performed parameter tuning on the same tuning set and reported the best configuration.

In contrast, our approach explicitly accumulates geometric distance and directional trends, achieving a tracking rate of 0.92, significantly outperforming all baselines. Our method does not require fixation or predefined goal transition probability and achieves the highest selection accuracy (0.97) with relatively small Min. Samples (3).

Statistical analyses are conducted on participant-level aggregated metrics. Since each of the data follows a repeated-measures design and the results are bounded and discretized, we use the Friedman test to compare the overall differences. Post-hoc pairwise comparisons were performed using the Wilcoxon signed-rank test with Holm correction. For selection accuracy and tracking rate, the Friedman test finds significant differences between all groups ($p<0.05$). Wilcoxon signed-rank test shows that our method yields a significantly higher tracking rate (min. difference = 0.11, $p<0.05$) and higher selection accuracy (min. difference = 0.09, $p<0.005$).

 
\begin{table}[t]
\vspace{-10pt}
\setlength{\tabcolsep}{1pt} 
\caption{Intent Recognition Robustness}
\vspace{-10pt}
\label{intent}
\begin{center}
    \centering
    \begin{tabular}{ccccc}
     \hline
     
     \multicolumn{1}{c}{\multirow{2}{*}{\shortstack{\\Methods}}} & \multicolumn{1}{c}{\multirow{1}{*}{\shortstack{Tracking Rate}}}& \multicolumn{1}{c}{\multirow{1}{*}{\shortstack{Selection Accuracy }}} & \multicolumn{1}{c}{\multirow{1}{*}{\shortstack{Min. Samples }}} 
     \\ & (Dynamic) & (Static) & (Static)\\ \hline
     kNN-based \cite{11037823}  & 0.21$\pm$0.15& 0.63$\pm$0.14 & 1 \\ 
    Fixation-based  \cite{tay2025intent} &0.13$\pm$0.12 &0.84$\pm$0.10 & 5\\
    Distribution-based \cite{10806588}& 0.72$\pm$0.11& 0.90$\pm$0.07& 3
     \\ 
     HMM-based \cite{8593766} & 0.88$\pm$0.09 & 0.78$\pm$0.12 &  20\\
   LSTM-based \cite{gonzalez2019perceptually}& 0.28$\pm$0.11 & 0.73$\pm$0.11 & 25 \\
   Our
  &\textbf{0.92$\pm$0.08} &\textbf{0.97$\pm$0.03} & 3\\ 
  
    \hline
    \end{tabular}
    \end{center}
    \footnotesize{Min. Samples: the minimum number of sampling points required to recognize intent. HMM-based method needs a temporal average filter, and the LSTM-based method can only run with a fixed-length time window, resulting in a large number of sampling points being required.  }
    \vspace{-5
    pt}
\end{table}

\begin{table}[tb]
\setlength{\tabcolsep}{1pt} 
\caption{Multi-Perspective Alignment Robustness}
\label{multi-perspective}
\begin{center}
    \centering
    \begin{tabular}{cccccccccc}
     \hline
     
     \multicolumn{1}{c}{\multirow{2}{*}{\shortstack{\\Distance (cm)}}} & \multicolumn{3}{c}{\shortstack{\\ArUco \cite{9889538} }}& \multicolumn{3}{c}{\shortstack{\\Matching \cite{lai2025famhrifoundationmodelassistedmultimodal}}}& \multicolumn{3}{c}{\shortstack{\\Our }} \\ \cmidrule(r){2-4} \cmidrule(r){5-7} \cmidrule(r){8-10} 
     &  $0^\circ$   & $90^\circ$ & $180^\circ$  &  $0^\circ$   & $90^\circ$ & $180^\circ$  &  $0^\circ$   & $90^\circ$ & $180^\circ$
     \\ \hline
    30 & \textbf{100} &\textbf{100}& \textbf{100}
& \textbf{100} &61 & 57
& \textbf{100} & \textbf{100} & \textbf{100}   \\ 
   40 & 99 & \textbf{100} & \textbf{100}
& \textbf{100} & 74 & 62
& \textbf{100} & \textbf{100} & \textbf{100} \\
  50 & 78 & 72 & 75
& 84 & 53 & 46
& \textbf{98} & \textbf{96} & \textbf{99}\\
   60 & 48 & 52 & 61
& 19 & 21 & 17
& \textbf{94} & \textbf{91} & \textbf{92}
    \\ 
 70
   & 41 & 38 & 32
& 13 & 9 & 11
& \textbf{84} & \textbf{86} & \textbf{89}\\ 
  
    \hline
    \end{tabular}
    \end{center}
    \footnotesize{ }
    \vspace{-10pt}
\end{table}
\vspace{-5pt}
\subsection{Multi-Perspective Alignment Robustness}

Tab. \ref{multi-perspective} compares the number of successful trials. At short distances (30–40 cm), all methods achieve high robustness. However, as distance increases, ArUco-based \cite{9889538} alignment degrades due to marker detection loss, and feature-matching methods \cite{lai2025famhrifoundationmodelassistedmultimodal} fail to reliably match features on objects.

In contrast, our alignment method remains stable across all tested distances and angles, maintaining 84 successful trials even at 70 cm. This demonstrates strong robustness to viewpoint variation and long-range observation.

\begin{table*}[htp]
\vspace{5pt}
\setlength{\tabcolsep}{2pt} 
\caption{Comparison of Task Completeness across Different Gaze-based Robot Systems}
\vspace{-9pt}
\label{task}
\begin{center}
    \centering
    \begin{tabular}{ccccccccccc}
     \hline
\multicolumn{1}{c}{\multirow{2}{*}{\shortstack{\\Methods}}}& \multicolumn{1}{c}{\multirow{2}{*}{\shortstack{\\Multi \\Modaility?}}} &\multicolumn{1}{c}{\multirow{2}{*}{\shortstack{\\Real-time\\ Feedback?}}} & \multicolumn{1}{c}{\multirow{2}{*}{\shortstack{\\ Fixation and\\
GUI Free?}}} &\multicolumn{1}{c}{\multirow{2}{*}{\shortstack{\\Handles \\
overlapping?}}} & \multicolumn{2}{c}{\shortstack{\\Success rate}} &  \multicolumn{2}{c}{\shortstack{\\Command Duration (s)}} & \multicolumn{2}{c}{\shortstack{\\Task Duration (s)}}   \\ \cmidrule(r){6-7}   \cmidrule(r){8-9}   \cmidrule(r){10-11}  
      &&&&& $S_3$ & $S_4$ & $S_3$ & $S_4$ & $S_3$ & $S_4$      \\ \hline
    MR-GUI \cite{c7} &\textcolor{red}{\ding{55}} &\textcolor{red}{\ding{55}}&\textcolor{red}{\ding{55}}& \textcolor{red}{\ding{55}}&0.93$\pm$0.09  &0.61$\pm$0.07 & 18.3$\pm$3.8& 5.8$\pm$1.2& 52.6$\pm$5.9&34.7$\pm$2.3\\
    FreeView \cite{c24} &\textcolor{red}{\ding{55}} &\textcolor{green}{\ding{51}}(teleoperation) &\textcolor{red}{\ding{55}} &\textcolor{red}{\ding{55}}&- & -& 193.8$\pm$36.2 & 167.9$\pm$33.7 &- &- \\
    GlanceGaze \cite{tay2025intent} &\textcolor{red}{\ding{55}} &\textcolor{red}{\ding{55}} & \textcolor{red}{\ding{55}} & \textcolor{red}{\ding{55}}& 0.67$\pm$0.14 & 0.59$\pm$0.17 & 9.6$\pm$1.7 & 4.4$\pm$0.9  & 48.7$\pm$4.5 & 36.1$\pm$4.3\\
    FAM-HRI \cite{lai2025famhrifoundationmodelassistedmultimodal} &\textcolor{red}{\ding{55}}& \textcolor{red}{\ding{55}}& \textcolor{green}{\ding{51}}& \textcolor{red}{\ding{55}}& 0.93 $\pm$ 0.07 & 0.73$\pm$0.08 &8.4$\pm$0.71&2.9$\pm$0.31& 41.6$\pm$5.3& 32.4$\pm$3.9 \\ 
   Our  &\textcolor{green}{\ding{51}}&\textcolor{green}{\ding{51}} &\textcolor{green}{\ding{51}} & \textcolor{green}{\ding{51}}
   & \textbf{0.98$\pm$0.04} & \textbf{0.91$\pm$0.05} & \textbf{5.2$\pm$0.52} & \textbf{2.4$\pm$0.18} & \textbf{36.4$\pm$2.2} & \textbf{29.5$\pm$1.8}\\ 
  
    \hline
    \end{tabular}
    \end{center}
    \vspace{-4pt}
    \footnotesize{To eliminate the influence of different action generators, all baseline methods except the FreeView approach employ similar behavior tree-based planners. The FreeView method is based on teleoperation; therefore, the success rate is meaningless. Additionally, task duration equates to command duration. No confirmation process is performed in this experiment.}
    \vspace{-9pt}
\end{table*}

\subsection{Task Completeness}
We compare our system with representative gaze-based HRI pipelines, as shown in Tab.~\ref{task}. MR-GUI~\cite{c7} and GlanceGaze~\cite{tay2025intent} rely on fixation-based object selection. MR-GUI requires fixation on a GUI to choose actions, increasing command duration (18.3s \& 5.8s). GlanceGaze infers actions using a VLM, but depends on repeated fixation for object selection. Repeated fixations lengthen command duration (4.4s) and reduce success rate (0.59) in $S_4$, where object overlap makes accurate gaze-based selection difficult. Moreover, GlanceGaze occasionally misinterprets action intent, e.g., confusing put and pour, resulting in lower success rates (0.67) in $S_3$. FreeView~\cite{c24} performs teleoperation through a virtual GUI around the robot end-effector. This requires continuous manual control, leading to extremely long command duration (193.8s \& 167.9s) and reduced practicality. FAM-HRI~\cite{lai2025famhrifoundationmodelassistedmultimodal} combines weighted gaze clustering with speech for multimodal interaction, reducing command duration compared with unimodal interaction. However, its intent inference is based on aggregated gaze observations and is not designed for continuous object-level intent inference. Since it does not continuously model the geometric relationship between gaze motion and object regions, this also leads FAM-HRI to follow the discrete target-pose triggering paradigm, in which the robot remains stationary until the target is determined. Therefore, the user receives no real-time feedback during command formation and cannot anticipate an incorrect target selection before execution is triggered. These limitations become more evident in $S_4$, where overlapping objects and viewpoint-induced occlusion make gaze-based object grounding more ambiguous. In contrast, our pipeline continuously updates object-centric intent confidence, allowing the continuous shared controller to move toward a confidence-weighted virtual target before final command confirmation, reducing task duration. This pre-command motion provides immediate feedback about the current inferred target, enabling earlier correction. 




\subsection{User Study}

\begin{figure}[t]
      \centering
      \vspace{-3pt}
      \includegraphics[width=0.42\textwidth]{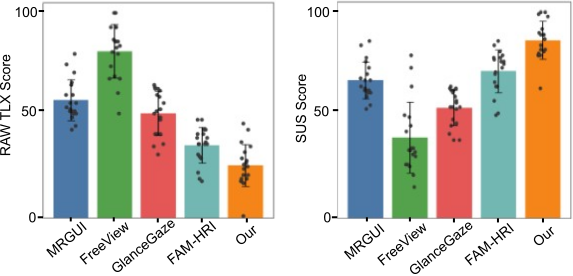} 
     \vspace{-5pt}
      \caption{Final user outcomes for the raw NASA-TLX (right) and System Usability Score (left).}
      \vspace{-5pt}
      \label{user}
\end{figure}

After completing all baseline and proposed methods, participants completed a NASA-TLX workload assessment 
and a System Usability Scale (SUS). The distribution and summary statistics are shown in Fig. \ref{user} and Tab. \ref{resultuser}.  

\begin{table}[tp] 
\vspace{-5pt}
\caption{Results of User Study}
\vspace{-9pt}
\label{resultuser}
\begin{center}
    \centering
    \begin{tabular}{ccc}
     \hline
    Methods &  NASA-TLX $\downarrow $ & SUS $\uparrow$\\ \hline
     MR-GUI \cite{c7}  & 57.42$\pm$ 9.96& 66.89$\pm$9.05 \\ 
    FreeView \cite{c24}  &81.21$\pm$13.14 &39.05$\pm$17.31\\
     GlanceGaze  \cite{tay2025intent}  &51.05$\pm$10.91 &53.57$\pm$8.61\\
      FAM-HRI \cite{lai2025famhrifoundationmodelassistedmultimodal}  &35.36$\pm$8.71 &71.36$\pm$10.32\\
    Our  & \textbf{25.57$\pm$11.34}&\textbf{86.42$\pm$9.29} \\
    \hline
    \end{tabular}
    \end{center}
   \vspace{-15pt}
\end{table}

For NASA-TLX results, a Shapiro–Wilk test confirms that all groups followed a normal distribution ($p>0.05$).
ANOVA results show significant differences between the groups ($p<0.05$). Post-hoc Tukey HSD analysis shows that both FAM-HRI and our method yield significantly lower cognitive load than other baselines (minimum difference = $15.68$, 
$p<0.05$), and our method further achieves significantly lower cognitive load than FAM-HRI (difference = $9.78$, $p<0.05$). These results indicate that incorporating speech reduces cognitive load by explicit action intent, while the sticky-glance algorithm further lowers cognitive load by eliminating the need for fixation with a short glance.

For SUS results, a Shapiro–Wilk test confirms that all groups followed a normal distribution ($p>0.05$) except MR-GUI ($p=0.03<0.05$). ANOVA results show significant differences between the groups ($p<0.05$). Post-hoc Tukey HSD analysis shows that our method had significantly higher scores than other baselines (minimum difference = $15.05$, $p<0.05$). Participants reported that our interaction paradigm reduces learning effort and improves task success, leading to a clear user preference for the proposed approach.

\subsection{Ablation Studies}
\subsubsection{Continuous Shared  Control}
 
\begin{figure}[t]
      \centering
      \vspace{-5pt}
      \includegraphics[width=0.37\textwidth]{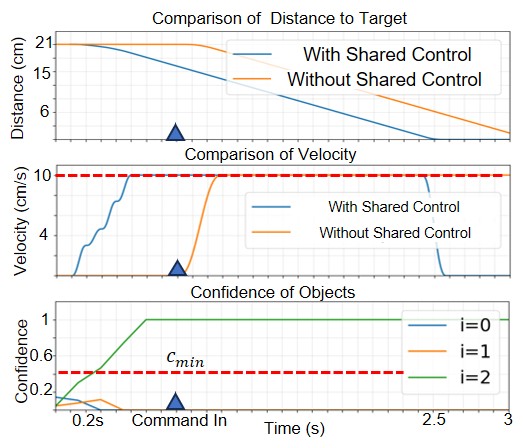} 
     \vspace{-10pt}
      \caption{Ablation Study of Continuous Shared Control}
      \label{jointly}
\end{figure}

We compare the proposed shared control to target-pose control by analyzing end-effector velocity and distance to the target. As shown in Fig. \ref{jointly}, without shared control, the robot remains stationary until an explicit command is given. In contrast, shared control enables continuous pre-command motion toward a potential virtual target, reducing the remaining distance by nearly 40\% once the command is explicit.

\subsubsection{Sticky-Glance Algorithm}

Tab. \ref{ablationsticky} evaluates the contribution of $e_i^{dist}$ and  $e_i^{dir}$. Removing any evidence reduces robustness under dynamic motion and static target.  

The full model achieves the highest tracking rates across all settings (0.92 under dynamic and 0.97 under static), demonstrating that jointly modeling distance and motion direction is essential for stable intent grounding.

\begin{table}[t]

\caption{Ablation Study of Sticky-Glance Algorithm}
\label{ablationsticky}
\begin{center}
    \centering
    \begin{tabular}{ccc}
     \hline
     
      \multicolumn{1}{c}{\multirow{2}{*}{\shortstack{\\Methods}}} & \multicolumn{1}{c}{\multirow{1}{*}{\shortstack{Tracking Rate}}}& \multicolumn{1}{c}{\multirow{1}{*}{\shortstack{Selection Accuracy }}} 
     \\ & (Dynamic) & (Static) \\ \hline
      w.o. $e_i^{dir}$ &0.86$\pm$0.19& 0.93$\pm$0.08\\
   w.o. $e_i^{dist}$ & 0.74$\pm$0.13 & 0.81$\pm$0.13\\
  our  &\textbf{0.92$\pm$0.08} &\textbf{0.97$\pm$0.03} \\
    \hline
    \end{tabular}
    \end{center}
    \footnotesize{ }
    \vspace{-5pt}
\end{table}

\begin{table}[t] 
\vspace{-10pt}
\caption{Ablation study of Different modalities}
\label{modality}
\begin{center}
    \centering
    \vspace{-12pt}
    \begin{tabular}{ccc}
     \hline
    Modality & Command Duration (s)& Success rate \\ \hline
     w.o. Gaze  & 4.48$\pm$0.89&0.79$\pm$0.18\\ 
    w.o. Speech  &3.93$\pm$0.72&0.87$\pm$0.08\\
    Our  & \textbf{2.16$\pm$0.16}&\textbf{0.93$\pm$0.04 } \\
    \hline
    \end{tabular}
    \end{center}
    \vspace{-18pt}
\end{table}

\subsubsection{Interaction Modalities}
We evaluate different interaction modalities in $S_4$ (100 trials). In the test area with  3×3 Duplo blocks, the user is required to grasp a specified block.

As shown in Tab. \ref{modality}, speech interaction (w.o. Gaze) leads to longer command duration and lower success rates. Commands such as \textit{“pick up the brick in the third row, second column}” are ambiguous and increase cognitive effort in spatial referencing. Gaze interaction (w.o. Speech) allows efficient object selection, but action selection requires looking at predefined ArUco markers as a GUI, which increases command duration and cognitive load \cite{11127558}. In contrast, our approach achieves the shortest command duration (2.16s) and highest success rate (0.93), demonstrating that combining gaze-based object grounding with speech-based action specification enables more efficient and reliable interaction.


\section{Conclusion and Future Work}
In this work, we presented Glance-Say, a multimodal human-robot interaction system that combines glance-based object grounding, speech-based action specification, and confidence-aware continuous shared control for assistive manipulation. The proposed sticky-glance algorithm converts sparse and noisy gaze signals into temporally stable, object-anchored intent, enabling reliable target selection from only a brief glance. By integrating confidence-weighted continuous shared control, the robot maintains readiness proactively while ensuring safe and smooth task execution. Extensive experiments demonstrate strong robustness under dynamic motion, multi-perspective alignment, and overlapping object scenarios. User studies further confirm reduced cognitive load and improved usability compared to representative gaze-based baselines. These results highlight that our system provides an efficient and reliable paradigm for HRI.
Despite the performance gains, our framework relies on some handcrafted components, which may limit its scalability and generalization across diverse, unstructured environments. Future work will focus on developing a unified end-to-end multimodal model to jointly learn gaze grounding, intent prediction, and action generation, thereby enhancing the system's adaptability in complex scenarios.

\bibliographystyle{IEEEtran}
\bibliography{IEEEexample}

\end{document}